\documentclass[sigconf, screen]{acmart}
\AtBeginDocument{%
  }

\setcopyright{acmcopyright}
\copyrightyear{2023}
\acmYear{2023}
\acmDOI{XXXXXXX.XXXXXXX}

\acmConference[ICMR '23]{International Conference on Multimedia Retrieval}{June 12--15, 2023}{Thessaloniki, Greece}
\acmPrice{15.00}
\acmISBN{978-1-4503-XXXX-X/18/06}

\acmSubmissionID{5248}

\begin{document}


\title{CNNs with Multi-Level Attention for Domain Generalization}


\author{Aristotelis Ballas}
\orcid{0000-0003-1683-8433}
\affiliation{%
	\institution{Harokopio University of Athens}
	\streetaddress{Omirou 9, Tavros}
	\city{Athens}
	\country{Greece}
	\postcode{177 78}
}
\email{aballas@hua.gr}

\author{Cristos Diou}
\orcid{0000-0002-2461-1928}
\affiliation{%
	\institution{Harokopio University of Athens}
	\streetaddress{Omirou 9, Tavros}
	\city{Athens}
	\country{Greece}}
\email{cdiou@hua.gr}

\renewcommand{\shortauthors}{Ballas et al.}

\begin{abstract}
  In the past decade, deep convolutional neural networks have achieved
  significant success in image classification and ranking and have therefore
  found numerous applications in multimedia content retrieval. Still, these
  models suffer from performance degradation when neural networks are
  tested on out-of-distribution scenarios or on data originating from previously unseen data \textit{Domains}. In the present work, we focus on this problem of Domain Generalization and propose an alternative neural network architecture for robust, out-of-distribution image classification. We attempt to produce a model that focuses on the causal features of the depicted class for robust image classification in the Domain Generalization setting. To achieve this, we propose attending to multiple-levels of information throughout a Convolutional Neural Network and leveraging the most important attributes of an image by employing trainable attention mechanisms. To validate our method, we evaluate our model on four widely accepted Domain Generalization benchmarks, on which our model is able to surpass previously reported baselines in three out of four datasets and achieve the second best score in the fourth one.
\end{abstract}



\received{13 February 2023}
\received[accepted]{1 April 2023}

\maketitle

\section{Introduction}
One of the most fundamental prerequisites for training robust and generalizable
machine learning (ML) models, is the ability to learn representations which
adequately encapsulate the underlying generating processes of a data
distribution \cite{bengio_representation_2013, scholkopf2022causality,
  9363924}. One way of approaching the above problem, is to guide a model to
learn \textit{disentangled} representations from the training data and uncover
the ones which remain \textit{invariant} \cite{arjovsky_invariant_2020} under
distribution shift. For example, a photograph of a dog shares similar traits
with an image of a cartoon dog, or even a sketch of a dog. A generalizable model
should be able to recognize the same class despite it being found in separate
\textit{Domains}.

\begin{figure}[h]
	\centering
	\includegraphics[width=\linewidth]{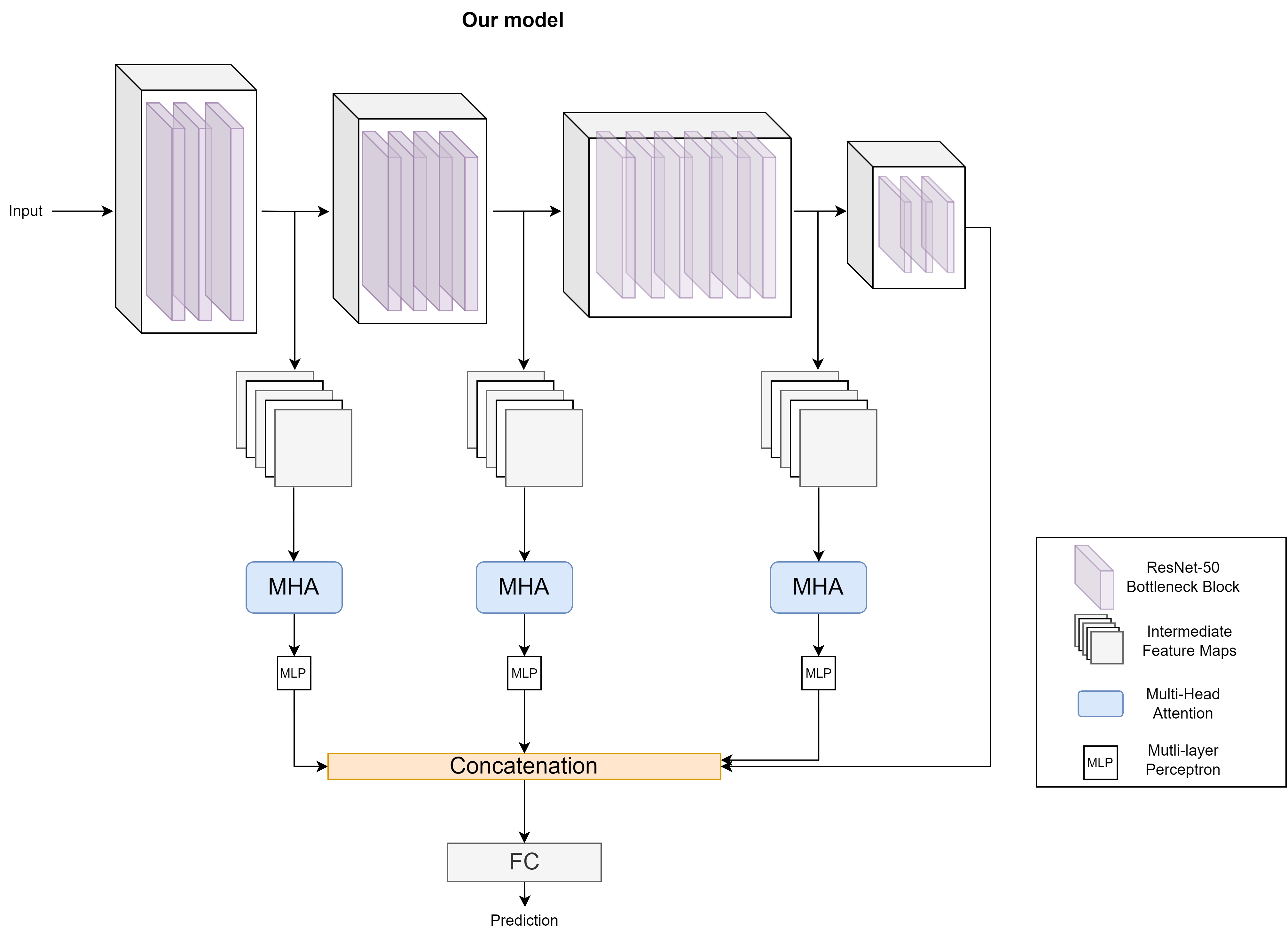}
	\caption{Visualization of our proposed framework on a ResNet-50 model. The
      feature maps after the 3rd, 7th and 13th bottleneck blocks are passed
      through multi-head attention layers with 3 heads each. Each re-weighted
      channel of the extracted feature maps is embedded into a vector of length
      32 via an Multi-Layer Perceptron and concatenated into a larger vector,
      along with the output of the last convolutional layer.}  
  	\Description{Our
      proposed framework.}
	\label{fig:model}
\end{figure}

ML models are trained under the assumption that the training and test data
distributions are independent and identically distributed. In practice however,
Deep Learning (DL) models are expected to mitigate, or not be affected by, the distribution shift between their training data and data they have not been presented with before. This is often not the case, as DL models often learn representations which entangle class-discriminative attributes with correlated, though irrelevant, features of images. They therefore fail to produce informative features and to generalize to unseen data domains
\cite{recht2019imagenet}. To this end, Domain Generalization (DG) \cite{zhou2022domainold, wang2022generalizing} methods aim at developing robust models which can generalize to unseen test data domains. Such methods attempt to address this problem by leveraging multiple domains in the training set, simulating biases found in real-world settings, synthesizing samples through augmentation and learning invariant representations through self-supervision (see Section \ref{sec:related_work}).

In 2017, Transformer networks \cite{vaswani_attention_2017} were proposed as a
 model for Natural Language Processing (NLP). Transformers introduced a
\textit{self-attention} mechanism for providing additional contextual
information to each word in an embedded sentence. Since their outstanding
success in several NLP tasks, Transformers and self-attention mechanisms have
slowly but steadily gained ground in the Computer Vision community
\cite{guo2022attention}, achieving significant advances in the field
\cite{hu2018squeeze, woo2018cbam, yang2020distilling}.
In this work, we argue that by attending to features extracted from multiple layers of a convolutional neural network via \textit{multi-head self-attention} mechanisms, a model can be trained to learn representations which reflect class-specific, domain-invariant attributes of an image. As a result, the trained model will be less affected by out-of-distribution data samples as it will base its predictions on the causal characteristics of the depicted class. Our contributions can be summarized in the following points:
\begin{itemize}
	\item We introduce a novel neural network architecture that utilizes
      self-attention \cite{vaswani_attention_2017} to attend to representations
      extracted throughout a Convolutional Neural Network (CNN), for robust out-of-distribution image classification in the DG setting
	\item We evaluate our proposed method on the widely adopted DG benchmarks of
      \cite{Li_2017_ICCV}, VLCS \cite{5995347}, Terra Incognita
      \cite{beery2018recognition} and Office-Home \cite{venkateswara2017deep}
      and demonstrate its effectiveness
	\item We provide qualitative visual results of our model's inference process
      and its ability to focus on the invariant and causal features of a
      class via saliency maps
\end{itemize}
In the next section we briefly present the most important contributions in DG, along with relative previous work in visual attention, from which we drew inspiration for our proposed algorithm.

\section{Related Work}
\label{sec:related_work}

\subsection{Domain Generalization}
There have been numerous efforts to address challenges related to domain shift
in the past (\cite{diou2010large}, \cite{wang2018deep}, \cite{weiss2016survey}),
however DG methods are different in that the model does not have \emph{any}
samples from the target domain(s) during training.

DG problems can be broadly categorized into two main settings, namely
\textit{multi-source} and \textit{single-source} DG \cite{zhou2022domainold}. In
multi-source DG, all algorithms assume that their training data originate from
$K$ (where $K > 1$) distinct but known data domains. These algorithms take
advantage of domain labels in order to discover invariant representations among
the separate marginal distributions. Most previously proposed methods fall
under this category. The authors of \cite{sun2016deep} propose deep CORAL, a
method which aligns the second-order statistics between source and target
domains in order to minimize the domain shift among their distributions. In
\cite{nam2021reducing}, Style-Agnostic networks, or SagNets, use an adversarial
learning paradigm to disentangle the style encodings of each domain and reduce
style-biased predictions. With a different approach, the authors of
\cite{zhou2021domain} investigate the usage of data augmentation and
style-mixing techniques for producing robust models. Another popular approach in
multi-source DG is Meta-learning, which focuses on learning the optimal
parameters for a source model from previous experiment
metadata. \cite{li2018learning, pmlr-v70-finn17a} and Adaptive Risk Minimization
(ARM) \cite{zhang2021adaptive}, all propose meta-learning algorithms for
adapting to unseen domains. Finally, \cite{10.1007/978-3-030-58607-2_12} uses
episodic training in the meta-learning setting to extract invariant
representations across source domains. On the other hand, single-source DG
methods hold no information about the presence of separate domains in their
training data, but assume that it originates from a single
distribution. Therefore, all single-source DG algorithms, such as our own,
operate in a domain-agnostic manner and do not take advantage of domain
labels. In \cite{Carlucci_2019_CVPR}, the authors combine self-supervised
learning with a jigsaw solving objective in order to reduce the model's
proneness to learning semantic features. Additionally, in \cite{Zhang_2021_CVPR}
the authors attempt to remove feature dependencies in their model via sample
weighting. Finally, RSC \cite{huangRSC2020} is a self-challenging training
heuristic to discard representations associated with very high gradients, which
forces the network to activate features correlated with the class and not the
domain.

\subsection{Visual Attention}

Attention mechanisms have long been introduced in CV
\cite{itti2001computational}, inspired by the human visual system's ability to
efficiently analyze complex scenes. More recently, attention mechanisms have
been proposed for the interpretation of the output of Convolutional Neural
Networks (CNNs), where they act as dynamic re-weighting processes which
\textit{attend} to the most important features of the input image. In
\cite{zhou2016learning}, the authors propose CAM, a post-hoc model
interpretation algorithm for estimating attention maps in classification
CNNs. Methods incorporating attention mechanisms into CNNs for image
classification have also been proposed in the past \cite{wang2017residual,
  chen2019all, ren2022shunted, yu2022metaformer}. In \cite{jetley2018learn}, the
authors introduce an end-to-end trainable mechanism for CNNs, by computing
compatibility scores between intermediate features of the network and a global
feature map. In \cite{woo2018cbam}, the Convolutional Block Attention Module, or CBAM, leverages both spatial and channel attention modules for adaptive feature refinement. Recently, several methods have been proposed which replace CNNs with \textit{self-attention} and \textit{multi-head attention} mechanisms
\cite{vaswani_attention_2017} applied directly on the image pixels
\cite{carion2020end, dosovitskiy2021an, liu2021swin}, leading to
transformer-based methods for CV \cite{han_survey_2023}.


\section{Methodology}
\begin{figure}[h]
	\centering
	\includegraphics[width=\linewidth]{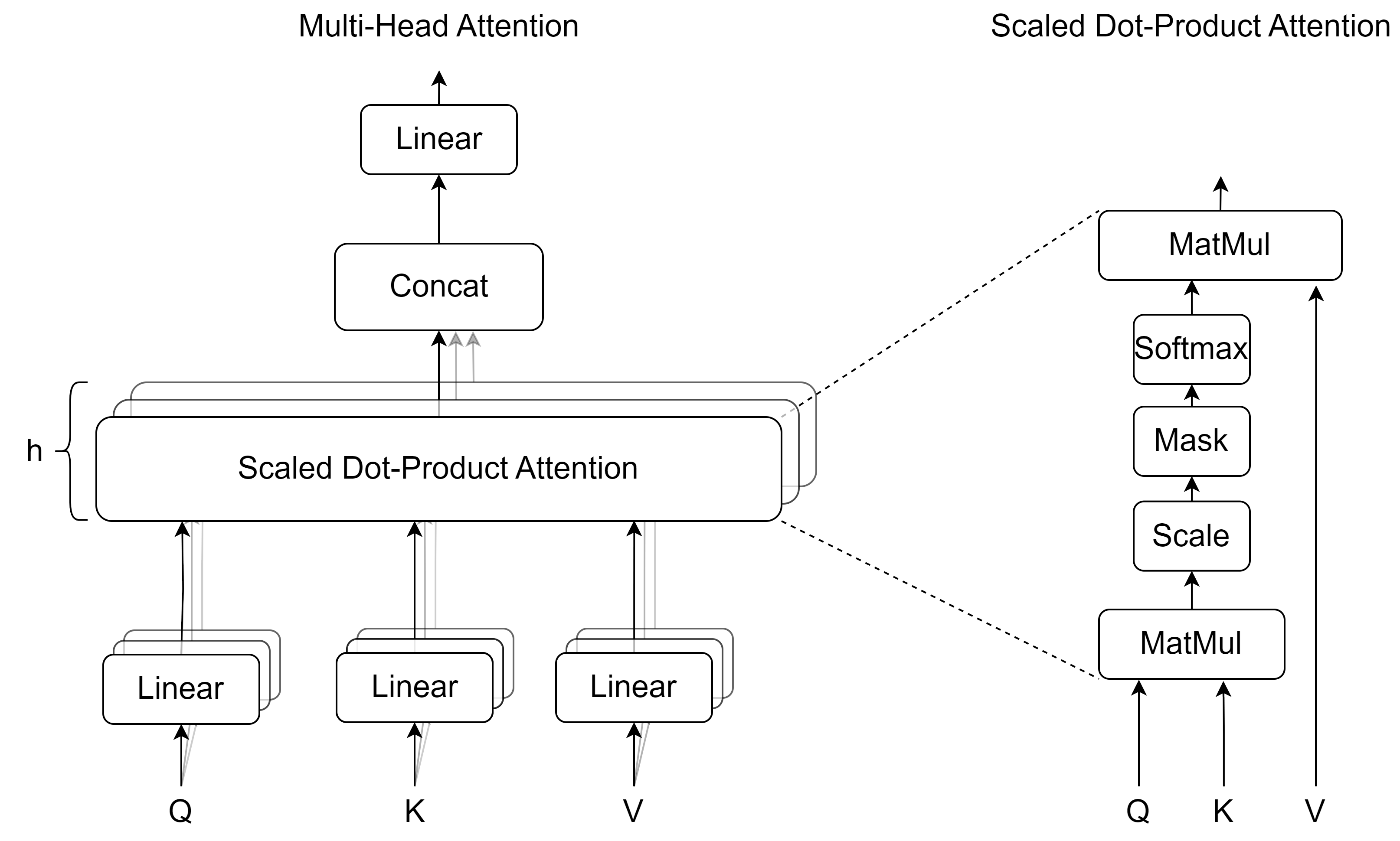}
	\caption{Visualization of the Multi-Head Attention mechanism. In our
      implementation, intermediate feature maps are extracted from a backbone
      ResNet-50 model and passed through a multi-head attention layer with 3
      heads ($h=3$). We propose attending to each channel of the extracted
      feature maps. For the compatibility metric in the self-attention module,
      we select to use the Scaled-Dot product.} 
	\Description{Visualization of the Multi-Head Attention}
	\label{fig:attention}
\end{figure}

Information passed through popular Convolutional Neural Network architectures,
such as ResNets \cite{He_2016_CVPR}, tends to get \textit{entangled} with
non-causal attributes of an image due to correlations in the data
distribution \cite{recht2019imagenet}. Our method is built around the hypothesis that this problem can be mitigated if we allow the network to select intermediate feature maps throughout a CNN for representation learning. We therefore extract feature maps at multiple network layers and pass them through a \textit{multi-head attention} mechanism (Figure \ref{fig:attention}). In our implementation we consider self dot-product attention with 3 heads. Given an intermediate feature map $\textbf{M} \in
\textbf{R}^{b \times c \times h \times w}$, where b is the batch size, c is the
number of channels and h and w are the height and width of the feature map, we
aim to attend to each of the channels. As a first step, we flatten the feature maps $\textbf{M}$ into a dimension of $(b, c, h \times w)$. We follow by linearly projecting the flattened feature maps into a $(b, c, d_{embed})$ dimension Tensor, where $d_{embed}$ is the size of each channel's embedded feature map. Each channel can be thought of as the token in the classic Transformer architecture. Given the embedded feature maps $\textbf{X} \in \textbf{R}^{b \times c \times d_{embed}}$ and trainable weight matrices $\textbf{W}^Q, \textbf{W}^K, \textbf{W}^V \in \textbf{R}^{d_{embed} \times d_k}$ ($d_k$ the inner self-attention layer dimension), we create the query, key and value vectors: $\textbf{Q} = \textbf{XW}^Q, \textbf{K} = \textbf{XW}^K, \textbf{V} = \textbf{XW}^V, \textbf{R}^{b \times c \times d_k}$, which are fed to the multi-head attention block. The self-attention layer is defined as:

\begin{equation}
	Attention(\textbf{Q, K, V}) = softmax(\frac{\textbf{Q}\textbf{K}^T}{\sqrt{d_k}}) \textbf{V}
\end{equation}

while the multi-head attention is:

\begin{equation}
	MultiHead(\textbf{Q, K, V}) = Concat(head_1, ..., head_h)\textbf{W}^O
\end{equation}

where:

\begin{equation}
	head_i = Attention(\textbf{QW}{_i^Q}, \textbf{KW}{_i^K}, \textbf{VW}{_i^V})
\end{equation}

After the extracted feature maps have been attended to and re-weighted, we pass
them through a Multi-Layer Perceptron (MLP) in order to allow our model to learn
a mapping between the processed features. The MLP consists of two Linear layers, activated by the GELU function \cite{hendrycks2016gaussian}. Finally, the projected features are flattened, concatenated and passed through a fully connected classification layer for the final decision. Our proposed framework is
visualized in Figure \ref{fig:model}.


\section{Experimental Setup}

In our experiments, we build our method on a vanilla ResNet-50
\cite{He_2016_CVPR} model, pre-trained on ImageNet. For our method, we choose to
extract intermediate feature maps from the 3rd, 7th and 13th bottleneck blocks
of the backbone ResNet-50 model, as shown in Fig \ref{fig:model}. We train our model with the SGD optimizer for 30 epochs and a batch size of 32 images. The learning rate is set at 0.001 and decays with a rate 0.1 at epoch 24. The proposed framework was implemented with the PyTorch library \cite{NEURIPS2019_9015} and trained on a NVIDIA RTX A5000 GPU.

We evaluate our method against 8 previous state-of-the-art algorithms, which use
a ResNet-50 as their base model. Specifically, the baseline models we select
are: ERM\cite{vapnik1999nature}, RSC \cite{huangRSC2020}, MIXUP
\cite{yan2020improve}, CORAL\cite{sun2016deep}, MMD \cite{li2018domain}, SagNet
\cite{nam2021reducing}, SelfReg \cite{kim_selfreg_2021} and ARM
\cite{zhang2021adaptive}. The above algorithms are a mix of both multi-source
and single-source methods allowing us to demonstrate the effectiveness of our
proposed method. The hyperparameters of each algorithm are set to reflect the ones in the original papers. All baselines are implemented and executed using the DomainBed \cite{gulrajani2021in} codebase for a fair comparison. The presented experimental results are averaged over 3 runs.

\subsection{Datasets}

To evaluate the robustness of our method we experiment on four well-known and
publicly available DG benchmark datasets, namely PACS \cite{Li_2017_ICCV}, VLCS
\cite{5995347}, TerraIncognita \cite{beery2018recognition} and Office-Home
\cite{venkateswara2017deep}. Specifically:
\begin{itemize}
	\item \textbf{PACS} contains images originating from the Photo, Art Painting, Cartoon and Sketch domains. It also contains a total of 9,991 images and 7 class labels.
	\item \textbf{VLCS} incorporates 10,729 real-world images from the PASCAL VOC, LabelMe, Caltech 101 and SUN09 datasets (or domains) and depicts 5 classes in total.
	\item \textbf{Terra Incognita} contains photographs of wild animals taken by trap cameras at 4 different locations (L100, L38, L43 and L46). This dataset contains 10 classes and 24,788 images in total.
	\item \textbf{Office-Home} comprises four domains of Art, Clipart, Product and Real-World images. The dataset contains 15,588 examples and 65 classes in total.
\end{itemize}
For each respective dataset we follow the standard \textit{leave-one-domain-out
  cross-validation} DG protocol, as described in \cite{Li_2017_ICCV,
  Ghifary_2015_ICCV}. In this setting, a target domain is selected and held out
from the model's training data split. The generalizability of the trained model
is then measured by its accuracy on the unseen data originating from the target
domain. For example, in the first experiment with the PACS dataset, the domains of Photo, Cartoon and Sketch are selected as \textit{Source} domains while the Art Painting domain is held out as the \textit{Target}. Therefore, the model is
trained on data from the source domains and evaluated on previously unseen art
images.

\subsection{Results}
The results of our experiments are presented in Table \ref{tab:results}. The
effectiveness of our method is demonstrated in the experimental outcome, as our
model is able to surpass previously proposed state-of-the-art algorithms in the
PACS, Terra Incognita and Office-Home datasets, while achieving the second best
performance in VLCS. In PACS, our model surpasses the previous best model by
1.06\%, while in TerraIncognita and Office-Home our implementation exceeds the
baselines by 0.98\% and 1.33\% respectively. What's more, even though our
algorithm is not able to achieve the top score in VLCS, it remains highly
competitive and ranks as second best among its predecessors.

\begin{table}\centering
	\begin{center}
		\caption{Top-1\% accuracy results, averaged over 3 runs, on the PACS, VLCS, Terra Incognita (denoted as Terra) and Office-Home (denoted as Office) datasets. The top results are highlighted in \textbf{bold} while the second best are underlined.}
		\label{tab:results}
		\begin{tabular}{|c||cccc|c|}
			\hline\noalign{}
		Method & PACS & VLCS & Terra & Office & Average \\
		\hline
		ERM    
		& 83.32 & 76.82 & 62.27 & 46.25 & 68.42 \\
		RSC    
		& 83.62 & 75.96	& 66.09	& 46.60	& 68.07 \\
		CORAL  
		& 83.57	& 76.97	& \underline{68.60}	& 48.05	& \underline{69.30} \\
		MIXUP   
		& 83.70 & \textbf{78.62} & 68.17 & 46.05 & 69.14 \\
		MMD    
		& 82.82	& 76.72 & 67.12	& 46.30	& 68.24 \\
		SagNet 
		& \underline{84.46} & 76.29 & 66.42	& \underline{48.60} & 68.94 \\
		SelfReg 
		& 84.16 & 75.46 & 65.73 & 47.00 & 68.09 \\
		ARM   
		& 83.78 & 76.38	& 63.49	& 45.50	& 67.29 \\
		\hline
		\textbf{Ours}
		& \textbf{85.52}& \underline{78.05} & \textbf{69.58} & \textbf{49.93} & \textbf{70.77} \\
		\hline
		
	\end{tabular}
\end{center}
\end{table}

\begin{figure}[h]
	\centering
	\includegraphics[width=\linewidth]{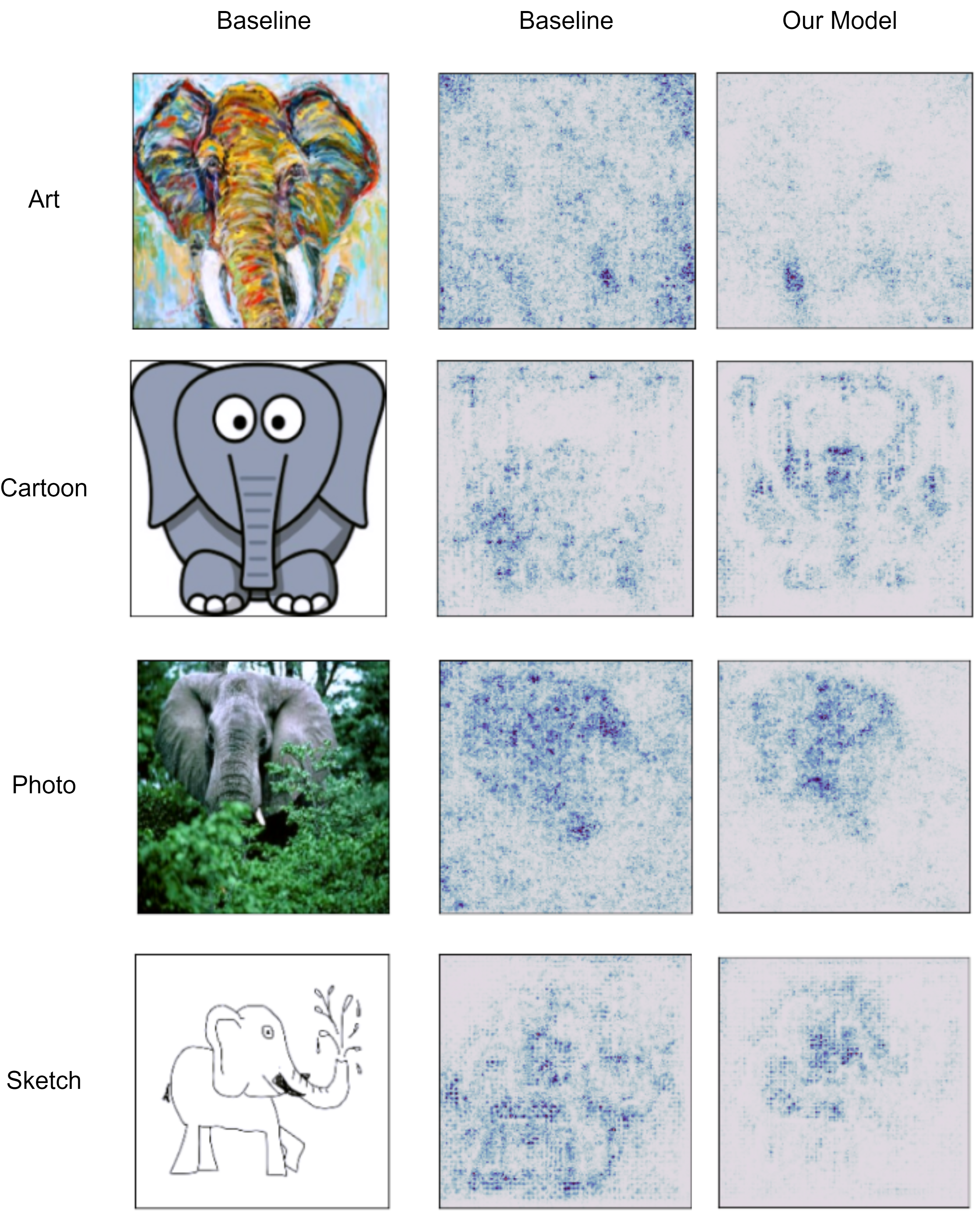}
	\caption{Saliency map visualization of the "elephant" class in 4 different
      domains, produced by a baseline ERM model and our method. The selected
      images are from the PACS dataset. By producing saliency maps, one is able
      to gain intuition into a model's inference process. The darker the pixel
      in the map, the more important it is to the model's prediction. The
      background noise and spurious correlations in the above images tend to
      contribute to the baseline model's decisions. In contrast, our framework 
      seems to pay attention to the invariant attributes of the class.}
	\Description{Saliency map visualizations}
	\label{fig:saliency}
\end{figure}

To further support our claims, we also provide visual examples of our
model's inference process via saliency maps. Specifically, we select to
implement the \textit{Image-Specific Class Saliency} method as proposed in
\cite{simonyan2014visualising}. In the above method, a visual map of the pixels
contributing the most to the model's prediction is produced by computing and
visualizing the gradient of the loss function with respect to the input
image. As depicted in Figure \ref{fig:saliency}, the darker a pixel, the more
significant it is to the model. We choose to visualize 4 images of the "elephant
class" from the four different domains in PACS. When compared to the baseline
ERM model, our method seems to base its decisions on features of the depicted
object (e.g. tusk of the elephant in the Art image) and pay less attention to
irrelevant attributes, such as the noisy backgrounds (e.g. tree leaves in the
Photo domain). This visual evidence proves promising towards researching
alternative architectures containing both convolutional and attention layers for
the DG setting. 

\section{Conclusions}
In this paper, we introduced a novel approach for image classification in the
Domain Generalization setting. The basic idea behind our implementation was to
allow the model to select the most class-discriminative and domain-invariant
representations via multi-head self-attention mechanisms which attend to
intermediate feature maps extracted from multiple layers of a convolutional neural network. The generalization ability of our model is supported
by extensive experimental results on four publicly available and well-known DG
benchmarks, in which our model either surpasses previously proposed algorithms or remains highly competitive. In addition, we provide visual qualitative examples of our model's inference process through saliency maps. The visual results demonstrate the fact that our model tends to disregard spurious correlations in its input images, such as background noise, and is able to base its predictions on class-specific attributes. However, our method still has room for improvement. The employment of multiple multi-head attention mechanisms and concatenation of embedded feature maps adds a significant computation and memory overhead, which is reflected by the relatively small image batch size in our experiments. For future work, we aim to further research the intersection between visual attention and fully convolutional networks in order to propose mechanisms which will be able to explicitly pay attention to the causal features of a class.

\begin{acks}
The work leading to these results has received funding from the European Union’s Horizon 2020 research and innovation programme under Grant Agreement No. 965231, project REBECCA (REsearch on BrEast Cancer induced chronic conditions supported by Causal Analysis of multi-source data).
\end{acks}

\bibliographystyle{ACM-Reference-Format}
\bibliography{references}

\end{document}